\documentclass[conference]{IEEEtran}
\IEEEoverridecommandlockouts
\usepackage{cite}
\usepackage{amsmath,amssymb,amsfonts}
\usepackage{algorithmic}
\usepackage{textcomp}
\usepackage{xcolor}

\usepackage{enumitem}
\usepackage{graphicx}
\usepackage{booktabs}
\usepackage{multirow}
\usepackage{url}
\usepackage{array}
\usepackage{pifont}
\usepackage{stfloats}
\usepackage{balance}
\usepackage{marvosym}

\newcommand{\cmark}{\ding{51}}  
\newcommand{\xmark}{\ding{55}}  

\def\BibTeX{{\rm B\kern-.05em{\sc i\kern-.025em b}\kern-.08em
    T\kern-.1667em\lower.7ex\hbox{E}\kern-.125emX}}
\begin{document}

\title{EVOLVE: Uncertainty-Aware Multivariate Time Series Anomaly Prediction with Adaptive Prototype Memories 
}

\author{
\IEEEauthorblockN{
Jianxin Jin\textsuperscript{1},
Kai Zhao\textsuperscript{2},
Shiyan Hu\textsuperscript{1},
Yang Shu\textsuperscript{1},
Peng Chen\textsuperscript{3},
Meng Wang\textsuperscript{3},
\\
Zhongwen Rao\textsuperscript{4},
Yijie Li\textsuperscript{5},
Jing Wang\textsuperscript{5},
Bin Yang\textsuperscript{1},
Chenjuan Guo\textsuperscript{1}
}

\IEEEauthorblockA{
\textsuperscript{1}\textit{East China Normal University},
\textsuperscript{2}\textit{Aalborg University},
\textsuperscript{3}\textit{2012 APPLab, Huawei},
\\
\textsuperscript{4}\textit{Huawei Technologies Co., Ltd},
\textsuperscript{5}\textit{Huawei Cloud, Huawei},
}

\IEEEauthorblockA{
\{jxjin,syhu\}@stu.ecnu.edu.cn,\{kaiz\}@cs.aau.dk,
\{yshu,byang,cjguo\}@dase.ecnu.edu.cn,
\\
\{chenpeng192,wangmeng71,raozhongwen,liyijie5,wangjing630\}@huawei.com
}

}

\maketitle

\begin{abstract}
Unsupervised multivariate time series anomaly prediction (AP) aims to estimate future anomaly risks from current observations, providing early-warning signals for proactive intervention and maintenance in complex systems. 
Existing unsupervised AP methods follow a perturbation-based paradigm, where predefined or random perturbations are injected into historical windows to simulate abnormal future evolution. However, these methods share perturbations across different variables and contexts, and learn deterministic history-to-future evolution. 
In reality, 
anomaly evolution is variable- and context-dependent: the same perturbation may be amplified into abnormal future dynamics in one variable or context, while being absorbed as normal fluctuation in another. 
In addition, future evolution is not uniquely determined by a historical window, since similar histories may lead to multiple plausible trajectories, such as  gradual degradation or abrupt changes. 
To address these problems, we propose \textbf{EVOLVE}, a unified framework that 
models unsupervised AP via variable- and context-conditioned perturbations and uncertainty-aware 
future reasoning. 
EVOLVE constructs context-specific anomaly perturbations through an adaptive anomaly prototype memory, which 
can dynamically update and retrieve relevant perturbations at the variable level. 
EVOLVE models uncertainty by sampling multiple latent historical states and decoding through complementary smooth and deviation-amplifying evolution hypotheses, and the disagreement among decoded future trajectories is used to estimate future anomaly risk. 
Extensive experiments on six real-world benchmark datasets demonstrate that EVOLVE consistently outperforms state-of-the-art anomaly prediction baselines.
\end{abstract}


\section{Introduction}

Traditional unsupervised time series anomaly detection (AD) aims to identify abnormal patterns after they have already occurred~\cite{AutoTSAD, zhang2023}. 
It has been widely studied due to its critical role in maintaining the safety and stability of complex systems, such as industrial control systems, cloud computing platforms, AIOps, and transportation systems~\cite{AD_good_use_1, AD_good_use_2}. 
However, in many real-world applications, detecting anomalies after their occurrence is often insufficient, since system failures or abnormal events may have already caused service degradation, financial loss, or safety risks. 
Time series anomaly prediction (AP) addresses this limitation by estimating the risk of future anomalies from current observations before abnormal events fully manifest~\cite{AP_after_time_1, AP_after_time_2}. 
By providing early-warning signals, unsupervised AP enables proactive intervention, predictive maintenance, and risk-aware decision making in complex systems. 
Despite its practical importance, unsupervised AP is challenging because future anomaly labels are typically unavailable during training, making it difficult to directly learn the transition from normal histories to abnormal futures.

To address the absence of future anomaly labels, existing unsupervised AP methods commonly construct surrogate supervision through perturbation-based training. 
Given historical observations that are assumed to be normal or mostly normal, these methods inject synthetic anomaly perturbations into historical windows and train the model to predict or distinguish the corresponding abnormal future evolution~\cite{existing_AP_methods_1, existing_AP_methods_2, existing_AP_methods_3, existing_AP_methods_4}. 
The perturbations are usually sampled from predefined anomaly libraries or generated by random noise, serving as proxy abnormal patterns when ground-truth anomaly evolution is unavailable~\cite{existing_AP_methods_2, temporal_patterns_1, AP_good_use_5}. 
This perturbation injection and evolution prediction paradigm provides a practical way to learn the transition from normal histories to potential abnormal futures, and has become a common framework for unsupervised anomaly prediction. 
Nevertheless, reliable AP in heterogeneous multivariate time series remains intrinsically challenging, because anomaly evolution is variable- and context-dependent, and future evolution is inherently uncertain.

\begin{figure}[t]
  \centering
  \includegraphics[width=\columnwidth]{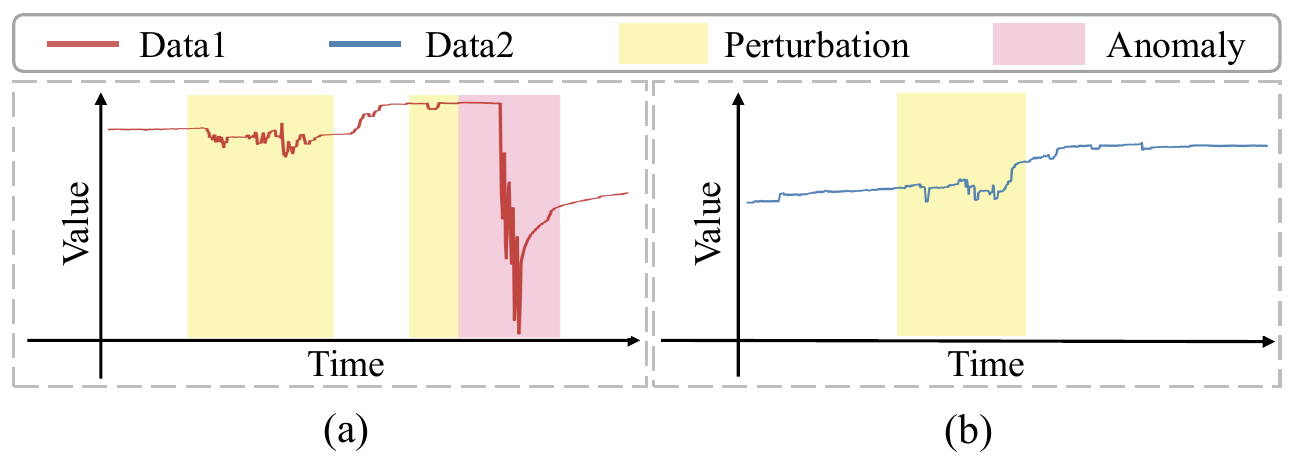}
    \caption{Different impacts of similar historical perturbations on future evolution under time series heterogeneity. 
  (a) Perturbations are amplified with abnormal evolution. 
  (b) Perturbations are absorbed with normal evolution.}
  \label{fig:introduction}
\end{figure}

\textbf{Challenge 1: Perturbation responses are highly heterogeneous across different variables and contexts~\cite{time_series_heterogeneous_1}.}
As illustrated in Figure~\ref{fig:introduction}, the same type of perturbation may be amplified into abnormal future evolution in one variable or context state, while being absorbed as a normal fluctuation in another. 
This phenomenon indicates that anomaly prediction cannot rely on globally shared or random perturbations; instead, the injected anomaly patterns should be conditioned on the variable-level characteristics and historical context of each time series. 
However, existing unsupervised anomaly prediction methods typically simulate abnormal futures using predefined anomaly libraries or random perturbations shared across different time series~\cite{existing_AP_methods_1, existing_AP_methods_2, existing_AP_methods_3}. 
Such static or context-agnostic perturbation strategies make it difficult to capture heterogeneous perturbation responses, especially when different variables exhibit distinct sensitivities to similar disturbances.

\textbf{Challenge 2: Future anomaly evolutions are highly uncertain.}
Anomaly prediction is inherently uncertain because the same historical window may correspond to multiple plausible future evolutions under different system conditions or failure modes.
On the one hand, noisy or incomplete historical observations may lead to ambiguous latent semantics, making it difficult to determine whether the current state is stable or approaching abnormal evolution~\cite{history_uncertainty_1}. 
On the other hand, even similar historical patterns may evolve into different future trajectories, such as normal absorption, gradual degradation, or abrupt abnormal changes~\cite{future_uncertainty_1, future_uncertainty_2}. 
Nevertheless, most existing anomaly prediction methods adopt deterministic mappings from history to future, producing a single predicted outcome or anomaly decision~\cite{existing_AP_methods_1, existing_AP_methods_4, existing_AP_methods_6}. 
This is insufficient to characterize the diversity of possible future evolutions, which leads to unreliable anomaly prediction. 

To address the above challenges, we propose \textbf{EVOLVE}, a unified framework for unsupervised multivariate time series anomaly prediction. 
Instead of treating perturbation injection as data augmentation, EVOLVE models anomaly prediction with \emph{variable-conditioned perturbation} and \emph{uncertainty-aware multi-hypothesis future reasoning}. 
For the first challenge, EVOLVE introduces an adaptive anomaly prototype memory to construct context-specific anomaly perturbations. 
The memory combines prior anomaly patterns with dynamically updated prototypes accumulated during training. 
Given a historical representation, EVOLVE retrieves relevant anomaly prototypes at the variable level and injects them into the latent representation, producing anomaly-based historical states tailored to the current time series. 
Moreover, an adversarial perturbation learning mechanism encourages the injection module to generate hard and informative perturbations, enabling the model to better distinguish heterogeneous anomaly responses across different variables and temporal contexts.

For the second challenge, EVOLVE develops an uncertainty-aware module together with a dual-structured decoder. 
The latent module parameterizes the distribution for historical representations and samples multiple possible states, capturing the uncertainty of historical semantics. 
Each sampled latent state is then decoded through two complementary evolution hypotheses: a smooth evolution hypothesis, which models normal absorption of perturbations, and a deviation-amplifying evolution hypothesis, which models the amplification of local deviations into potential abnormal dynamics.
By generating multiple plausible future paths and measuring their disagreement, EVOLVE explicitly characterizes predictive uncertainty and estimates future anomaly risk without relying on labeled anomalies.

The main contributions of this work are summarized as follows:
\begin{itemize}
    \item We formulate unsupervised time series anomaly prediction as variable-conditioned multi-hypothesis future risk estimation, where anomaly risk is inferred from the uncertainty among plausible future evolutions.
    
    \item We propose an adaptive anomaly prototype memory with variable-level perturbation retrieval and dynamic updating, enabling context-specific anomaly injection beyond static or globally shared perturbation libraries.
    
    \item We design an uncertainty-aware latent modeling and dual-structured decoding mechanism to generate multiple future evolution paths under smooth and deviation-amplifying hypotheses. 
    
    \item Extensive experiments on six real-world benchmark datasets demonstrate the effectiveness and efficiency of EVOLVE. 
\end{itemize}

\section{RELATED WORK}

\subsection{Unsupervised Time Series Anomaly Prediction}

Time series anomaly prediction aims to anticipate future abnormal events before they fully occur, providing more proactive warning signals than conventional anomaly detection. 
Compared with post-hoc anomaly detection, unsupervised anomaly prediction is more challenging because future anomaly labels are unavailable during training, making it difficult to directly learn the transition from normal histories to abnormal futures~\cite{AP_good_use_4, AP_good_use_5, AP_good_use_6, existing_AP_methods_1, existing_AP_methods_2}.

One line of studies formulates anomaly prediction as precursor identification.
PAD introduces Precursor-of-Anomaly (PoA) detection to identify early signals that precede future anomalies~\cite{AP_after_time_4}. 
IGCL develops a generative contrastive learning framework that constructs diverse anomaly precursor samples without requiring anomaly labels~\cite{existing_AP_methods_2}. 
TAP combines reconstructive learning with contrastive objectives to model anomaly-related patterns with different reaction times~\cite{existing_AP_methods_1}. 
These methods provide useful formulations for unsupervised AP, but their precursor or perturbation construction strategies are usually shared across different time series, making it difficult to adapt to heterogeneous variable-level responses.

Recent methods further model abnormal future evolution through forecasting-based objectives. 
A2P leverages anomaly-aware forecasting and synthetic anomaly prompting to inject anomaly-related representations during training~\cite{existing_AP_methods_3}. 
RED-F proposes a reconstruction-elimination based dual-stream contrastive forecasting framework, where a purified normal-pattern baseline and the original sequence are forecasted in parallel to amplify weak anomaly precursors through relative trajectory comparison~\cite{existing_AP_methods_4}. 
FCM introduces future context modeling for abnormality forecasting by exploiting future-oriented contextual information and normality correlations between observations and future contexts~\cite{existing_AP_methods_5}. 
These forecasting-based AP methods improve the modeling of future abnormal patterns and serve as strong baselines in our experiments. 
However, existing AP methods usually construct precursor or perturbation patterns in a globally shared manner, or model future evolution through deterministic forecasting objectives. 
Such context-agnostic perturbation construction and single-path future modeling are insufficient for heterogeneous multivariate time series, where the same disturbance may be amplified in one variable or temporal state but absorbed in another. 
In contrast, EVOLVE constructs anomaly perturbations conditioned on variable-level historical representations through an adaptive anomaly prototype memory, and further reasons over multiple plausible future evolutions for anomaly risk estimation.

\subsection{Point-level Anomaly Prediction and Future Uncertainty}

Existing anomaly prediction studies differ in the granularity of their prediction targets. 
Some methods adopt an interval-level formulation, where the objective is to determine whether an anomaly will occur within a future window based on historical observations~\cite{AP_after_time_4, existing_AP_methods_1, existing_AP_methods_2}. 
For example, PAD outputs an overall risk score for a future interval, while IGCL and TAP infer future anomaly likelihood by modeling precursor patterns or reaction periods~\cite{AP_after_time_4, existing_AP_methods_2, existing_AP_methods_1}. 
Although interval-level prediction is useful for early warning, it does not localize fine-grained anomalous timestamps within the future horizon, which limits its applicability in scenarios requiring precise intervention.

To provide more precise warning signals, recent studies have considered point-level anomaly prediction. 
You et al.~\cite{AP_good_use_6} formulate anomaly prediction as estimating future anomaly indicators at each timestamp. 
A2P, RED-F, and FCM further perform step-wise or future-oriented anomaly prediction for future sequences, enabling more fine-grained temporal localization~\cite{existing_AP_methods_3, existing_AP_methods_4, existing_AP_methods_5}. 
Our work also follows the point-level AP setting, where the model predicts anomaly risks for future timestamps rather than only producing an interval-level warning. 
However, point-level AP is inherently more difficult because future evolution before anomalies occur is often uncertain: similar historical windows may lead to normal absorption, gradual degradation, or abrupt abnormal changes.

Uncertainty-aware time series forecasting has been studied in probabilistic forecasting and VAE-style generative models~\cite{future_uncertainty_2, hybrid}, which usually aim to model the distribution of future observations or improve forecasting likelihood. 
However, these methods are not specifically designed for anomaly prediction, where the key objective is to estimate whether future trajectories will diverge toward abnormal states. 
Most existing AP methods still adopt deterministic history-to-future mappings and generate a single future outcome, limiting their ability to characterize multiple plausible future evolutions. 
Different from generic stochastic forecasting, EVOLVE performs task-specific uncertainty modeling for AP: it samples multiple latent historical hypotheses and decodes them through smooth and deviation-amplifying evolution hypotheses. 
The predictive disagreement among decoded future trajectories is then used to estimate future anomaly risk.

\section{Methodology}
\subsection{Problem Setup}
\label{sec:problem}

Let $X_{1:T}=(x_1,x_2,\dots,x_T)\in\mathbb{R}^{T\times C}$ denote a historical multivariate time series window of length $T$ with $C$ variables, where $x_t\in\mathbb{R}^{C}$ is the observation at time step $t$. 
Given $X_{1:T}$, unsupervised time series anomaly prediction (AP) aims to estimate anomaly risks for the future horizon $X_{T+1:T+F}$ and produce future anomaly labels $\hat{Y}_{T+1:T+F}=(\hat{y}_{T+1},\dots,\hat{y}_{T+F})$, where $\hat{y}_t\in\{0,1\}$ indicates whether time step $t$ is predicted as anomalous~\cite{existing_AP_methods_1, existing_AP_methods_2}. 
During training, future observations can be used as self-supervised signals to learn future evolution, while future anomaly labels are unavailable. 
During inference, the true future observations are unavailable, and the model must estimate future anomaly risks solely from the historical window. 

\begin{figure*}[t]
  \centering
  \includegraphics[width=\textwidth]{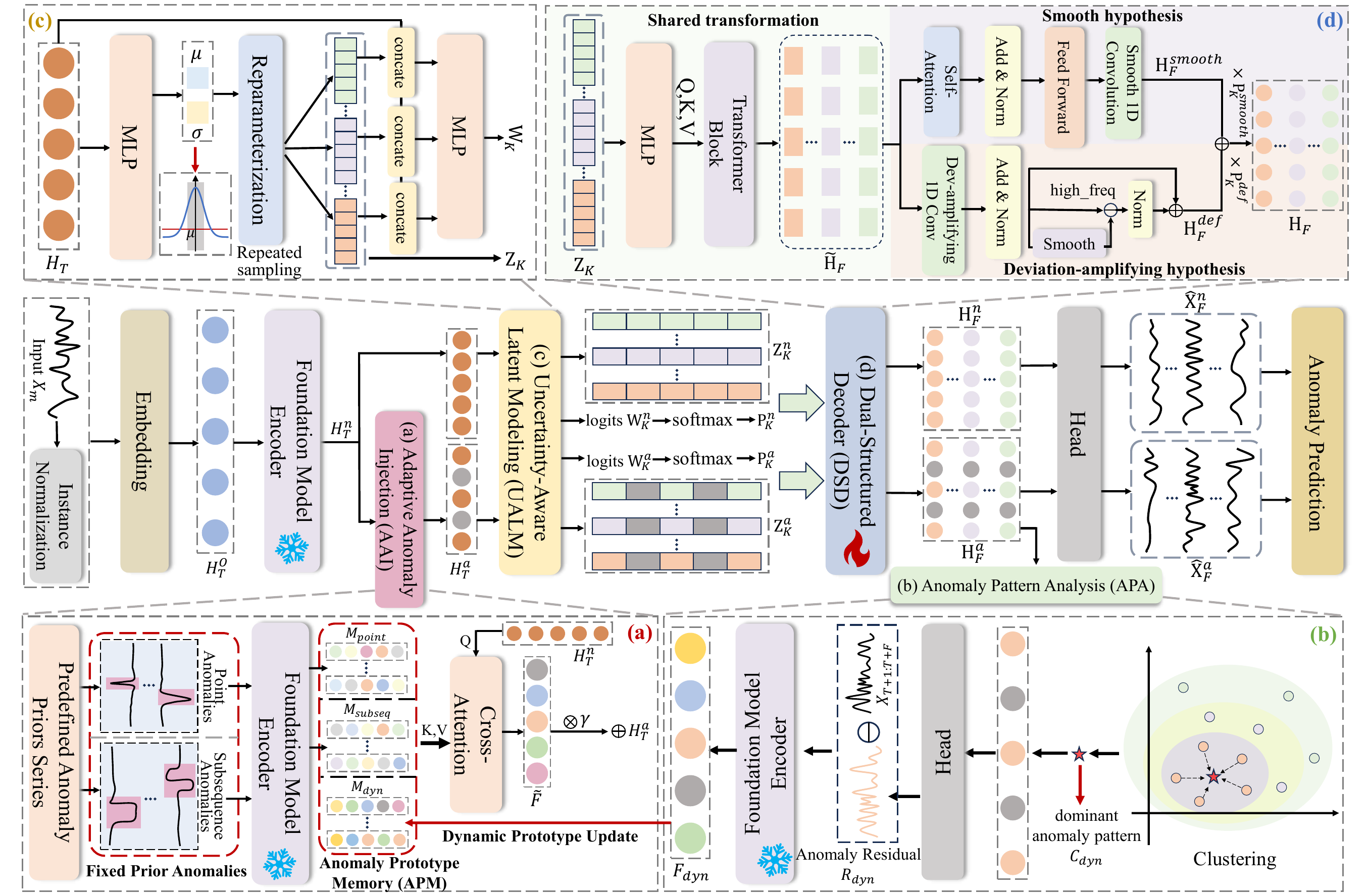}
  \caption{
    Overall architecture of EVOLVE. 
    Given a historical window, EVOLVE first encodes the input into a normal historical representation $H_T^n$. 
    During training, Adaptive Anomaly Injection (AAI) retrieves variable-conditioned perturbations from the Anomaly Prototype Memory (APM) and constructs an anomaly-aware historical representation $H_T^a$, while Anomaly Pattern Analysis (APA) progressively updates the dynamic prototype bank using residual-based anomaly-aware future representations. 
    Both the normal and anomaly-aware historical views are then processed by Uncertainty-Aware Latent Modeling (UALM), which samples multiple latent semantic hypotheses and estimates their evolution-mode probabilities. 
    The Dual-Structured Decoder (DSD) decodes these hypotheses through smooth and deviation-amplifying evolution hypotheses to generate multiple plausible future trajectories. 
    During inference, anomaly-aware injection, APA, and dynamic prototype updating are disabled, and future anomaly scores are computed from the predictive disagreement among decoded trajectories.
  }
  \label{fig:overall_architecture}
\end{figure*}

In EVOLVE, anomaly risk is estimated from the predictive disagreement among multiple decoded future trajectories. 
Given a historical window, the model generates $K$ plausible future trajectories:
\begin{equation}
    \hat{X}_{T+1:T+F}^{(k)} =
    (\hat{x}_{T+1}^{(k)}, \hat{x}_{T+2}^{(k)}, \dots, \hat{x}_{T+F}^{(k)}),
    \quad k=1,\dots,K,
\end{equation}
where $\hat{x}_{t}^{(k)}\in\mathbb{R}^{C}$ denotes the $k$-th predicted future observation at time step $t$. 
We emphasize that the anomaly score is not computed from the variance of latent variables, but from the predictive variance among multiple decoded future trajectories. 
Intuitively, if the historical window admits a stable future evolution, different plausible trajectories should remain consistent. 
In contrast, when the system is approaching abnormal evolution, plausible future trajectories tend to diverge, resulting in higher predictive disagreement.

Specifically, let $\hat{x}_{t,c}^{(k)}$ denote the prediction of the $c$-th variable at future time step $t$ from the $k$-th trajectory. 
The anomaly score at time step $t$ is computed by aggregating variable-wise predictive variance:
\begin{equation}
    s_t =
    \frac{1}{C}\sum_{c=1}^{C}
    \mathrm{Var}\left(
    \hat{x}_{t,c}^{(1)}, \hat{x}_{t,c}^{(2)}, \dots, \hat{x}_{t,c}^{(K)}
    \right),
    \quad t=T+1,\dots,T+F.
    \label{eq:anomaly_score}
\end{equation}
A larger $s_t$ indicates stronger disagreement among plausible future trajectories and thus higher future anomaly risk. 
Binary future anomaly labels are then obtained by thresholding the anomaly score:
\begin{equation}
    \hat{y}_t =
    \begin{cases}
    1, & s_t \ge \tau,\\
    0, & s_t < \tau,
    \end{cases}
    \quad t=T+1,\dots,T+F,
\end{equation}
where $\tau$ is the decision threshold.

\subsection{Design Principles}
\label{sec:design}

EVOLVE is built upon two design principles for unsupervised anomaly prediction in heterogeneous multivariate time series.

\textbf{Principle 1: Variable-conditioned perturbation modeling.}
Anomaly responses are variable- and context-dependent. 
The same perturbation may be amplified into abnormal future dynamics for one variable or context state, while being absorbed as normal fluctuation for another. 
Therefore, perturbations should not be treated as globally shared data augmentation patterns. 
Instead, they should be conditioned on variable-level historical representations and temporal contexts.

\textbf{Principle 2: Multi-hypothesis future reasoning.}
Future evolution in anomaly prediction is not uniquely determined by the historical window. 
Similar histories may lead to multiple plausible futures, including normal absorption, gradual degradation, and abrupt abnormal changes. 
Therefore, anomaly prediction should reason over multiple possible future trajectories rather than relying on a single deterministic prediction.

Following these principles, EVOLVE constructs variable-conditioned anomaly-aware historical views through adaptive anomaly prototype modeling, and estimates future anomaly risk from the disagreement among multiple decoded future trajectories.

\subsection{Overall Framework}
\label{sec:overall}

Figure~\ref{fig:overall_architecture} presents the overall framework of EVOLVE. 
The framework consists of three major components: 
(1) adaptive anomaly prototype modeling, which constructs an anomaly-aware historical view from the normal historical representation; 
(2) uncertainty-aware latent modeling, which samples multiple latent semantic hypotheses from historical representations; and 
(3) dual-structured decoding, which decodes each latent hypothesis into future trajectories under complementary evolution hypotheses.

Given an input historical window $X_{1:T}$, EVOLVE first applies instance normalization and an embedding layer to obtain an initial representation $H_T^O$. 
A time series encoder is then used to obtain the normal historical representation $H_T^n$. 
This representation serves as the \emph{normal historical view}, which describes the observed historical state without injected perturbations.

To model heterogeneous anomaly responses, EVOLVE further constructs an \emph{anomaly-aware historical view}. 
Specifically, the Adaptive Anomaly Injection (AAI) module retrieves relevant anomaly prototypes from the Anomaly Prototype Memory (APM) according to the current variable-level historical representation, and injects the retrieved perturbation into $H_T^n$ to obtain $H_T^a$. 
The APM contains both fixed prior anomaly patterns and dynamically updated anomaly prototypes. 
The dynamic prototype bank is initialized as an empty queue and is progressively updated during training by Anomaly Pattern Analysis (APA), rather than being randomly initialized. 
In this way, the anomaly-aware view provides context-specific perturbations for learning heterogeneous abnormal evolution.

Both the normal historical view $H_T^n$ and the anomaly-aware historical view $H_T^a$ are passed into the Uncertainty-Aware Latent Modeling (UALM) module. 
For each view, UALM parameterizes a latent distribution over historical semantics and samples $K$ latent hypotheses. 
Each latent hypothesis is also associated with a two-dimensional evolution-mode logit vector, corresponding to a smooth evolution hypothesis and a deviation-amplifying evolution hypothesis. 
Softmax is applied along this two-hypothesis dimension independently for each latent sample.

The sampled latent hypotheses are then decoded by the Dual-Structured Decoder (DSD). 
DSD contains a shared transformation module followed by two branches corresponding to complementary evolution hypotheses. 
The smooth evolution hypothesis models stable future evolution where perturbations are absorbed by the system, while the deviation-amplifying evolution hypothesis models the amplification of local deviations into potential abnormal dynamics. 
The two decoded representations are fused according to the evolution-mode probabilities, producing $K$ plausible future trajectories.

During training, both $H_T^n$ and $H_T^a$ are used. 
The normal historical view is encouraged to produce consistent future trajectories, while the anomaly-aware historical view is encouraged to produce diverse and deviation-aware trajectories. 
An adversarial perturbation learning strategy further encourages AAI to generate hard but plausible perturbations. 
During inference, anomaly-aware injection and dynamic prototype updating are disabled. 
EVOLVE only uses the normal historical view to generate $K$ future trajectories, and computes anomaly scores from their predictive variance as defined in Eq.~\eqref{eq:anomaly_score}.

\subsection{Adaptive Anomaly Prototype Modeling}
\label{sec:instance_modeling}

In heterogeneous multivariate time series, anomaly responses are highly variable- and context-dependent. 
A similar disturbance may be amplified into abnormal future dynamics for one variable or temporal state, while being absorbed as normal fluctuation for another. 
This motivates us to construct anomaly perturbations conditioned on the current historical representation, rather than using static or globally shared perturbations. 
To this end, EVOLVE introduces an adaptive anomaly prototype modeling mechanism, which constructs an anomaly-aware historical view $H_T^a$ from the normal historical view $H_T^n$.

This mechanism consists of three components. 
First, the Anomaly Prototype Memory (APM) stores both general anomaly priors and dynamically updated anomaly prototypes. 
Second, Adaptive Anomaly Injection (AAI) retrieves variable-relevant anomaly prototypes from APM and injects them into $H_T^n$ to obtain $H_T^a$. 
Third, Anomaly Pattern Analysis (APA) extracts residual-based dynamic prototypes from anomaly-aware future representations during training and updates the dynamic memory bank. 
The dynamic update is performed only during training, while inference uses only the normal historical view for future anomaly scoring.

\subsubsection{Anomaly Prototype Memory}

To support variable-conditioned perturbation modeling, we introduce the \emph{Anomaly Prototype Memory (APM)} as shown in Figure~\ref{fig:overall_architecture}(a). 
APM maintains three types of anomaly prototype banks: a point-wise prior bank, a subsequence-wise prior bank, and a dynamic prototype bank:
\begin{equation}
\mathcal{M}_{b} = \{F_i^{b}\}_{i=1}^{n_b}, 
\quad b \in \{\mathrm{point}, \mathrm{subseq}, \mathrm{dyn}\},
\end{equation}
where $F_i^b \in \mathbb{R}^{D}$ denotes an anomaly prototype in the same representation space as the historical representation, $n_b$ is the current number of stored prototypes, and $N_b$ is the capacity of the corresponding bank with $n_b \leq N_b$.

The point-wise and subsequence-wise banks provide general anomaly priors. 
They are initialized from predefined point-wise and subsequence-wise anomaly patterns~\cite{channel_independent_3, CrossAD}, which are encoded by the shared time series encoder to form fixed prior prototypes. 
These fixed banks offer stable references for common abnormal patterns, especially at the early stage of training.

In contrast, the dynamic prototype bank is designed to capture context-dependent anomaly patterns that cannot be fully covered by predefined priors. 
Importantly, the dynamic bank is initialized as an empty queue rather than random vectors, i.e., $n_{\mathrm{dyn}}=0$ at the beginning of training. 
At the beginning of training, prototype retrieval relies only on the fixed point-wise and subsequence-wise banks. 
As training proceeds, APA progressively writes residual-based dynamic prototypes into the dynamic bank, and a first-in-first-out (FIFO) strategy is adopted when the bank reaches its capacity. 
This design avoids introducing noisy random prototypes and enables the memory to gradually accumulate context-dependent anomaly patterns during training.

\subsubsection{Adaptive Anomaly Injection}

Given the normal historical view $H_T^n$, Adaptive Anomaly Injection (AAI) retrieves context-relevant anomaly prototypes from APM and constructs the anomaly-aware historical view $H_T^a$. 
Let
\begin{equation}
\mathcal{M}_{\mathrm{all}} = 
\mathrm{Concat}\left(
\mathcal{M}_{\mathrm{point}},
\mathcal{M}_{\mathrm{subseq}},
\mathcal{M}_{\mathrm{dyn}}
\right)
\end{equation}
denote the stacked memory of all available anomaly prototypes. 
AAI uses the normal historical representation as the query and the prototype memory as keys and values:
\begin{equation}
\tilde{F} = 
\mathrm{CrossAttn}
\left(
Q=H_T^n,
K=\mathcal{M}_{\mathrm{all}},
V=\mathcal{M}_{\mathrm{all}}
\right),
\end{equation}
where $\tilde{F}$ has the same representation shape as $H_T^n$ and denotes the retrieved anomaly perturbation conditioned on the current historical context.

The anomaly-aware historical view is then constructed as
\begin{equation}
H_T^a = H_T^n + \gamma \cdot \tilde{F},
\label{eq:adaptive_injection}
\end{equation}
where $\gamma$ is a learnable injection strength that controls the contribution of the retrieved perturbation. 
We use $\gamma$ to denote the injection strength and reserve the abnormal margin in the training objective as a separate symbol, avoiding notation ambiguity. 
Through cross-attention retrieval, AAI adaptively selects perturbations according to the current variable-level historical representation, enabling variable-conditioned anomaly injection rather than globally shared perturbation augmentation.

\subsubsection{Dynamic Prototype Updating}

The dynamic prototype bank is updated by \emph{Anomaly Pattern Analysis (APA)} during training, as shown in Figure~\ref{fig:overall_architecture}(b). 
The goal of APA is to refine the memory with residual patterns exposed by anomaly-aware decoding, so that APM can gradually adapt to heterogeneous and context-dependent anomaly evolution.

After the anomaly-aware historical view $H_T^a$ is constructed, the subsequent UALM and DSD modules decode it into $K$ anomaly-aware future representations. 
Let $\mathbf{H}_F^a=\{(H_F^a)^{(k)}\}_{k=1}^{K}$ denote these future representations. 
APA clusters these representations in the latent future space and selects the cluster with the largest number of members as the dominant anomaly-aware evolution pattern. 
The mean representation of this dominant cluster is denoted as
\begin{equation}
C_{\mathrm{dyn}}.
\end{equation}

Since $C_{\mathrm{dyn}}$ may still contain substantial normal future evolution context, directly writing it into the memory may blur anomaly-specific information. 
Therefore, APA first maps $C_{\mathrm{dyn}}$ to the time series space through a lightweight prediction head and computes its residual with respect to the observed future window:
\begin{equation}
R_{\mathrm{dyn}} = 
\mathrm{Head}(C_{\mathrm{dyn}}) - X_{T+1:T+F}.
\end{equation}
The residual highlights the deviation between anomaly-aware future evolution and the actual future observations, thereby suppressing normal background information and emphasizing anomaly-specific patterns.

The residual $R_{\mathrm{dyn}}$ is then encoded by the shared time series encoder to obtain a dynamic prototype:
\begin{equation}
F_{\mathrm{dyn}} = \mathrm{Encoder}(R_{\mathrm{dyn}}).
\end{equation}
Finally, $F_{\mathrm{dyn}}$ is written into the dynamic prototype bank. 
When the bank reaches its capacity, the oldest prototype is removed following FIFO replacement.

It is worth noting that APA is used only during training. 
Future observations are used as self-supervised signals to construct residual-based prototypes, similar to forecasting-based training objectives. 
During inference, the future window is unavailable, and both anomaly-aware injection and dynamic prototype updating are disabled. 
Therefore, no future information is used for test-time anomaly scoring.

\subsection{Uncertainty-Aware Multi-Hypothesis Future Modeling}
\label{sec:uncertainty_aware}

Future evolution in anomaly prediction is not uniquely determined by the historical window. 
Even similar histories may evolve into normal absorption, gradual degradation, or abrupt abnormal changes. 
Therefore, a single deterministic representation is insufficient to characterize future anomaly risk. 
To address this issue, EVOLVE introduces uncertainty-aware latent modeling and dual-structured decoding to generate multiple plausible future trajectories.

\subsubsection{Uncertainty-Aware Latent Modeling}

The \emph{Uncertainty-Aware Latent Modeling (UALM)} module models multiple semantic hypotheses from historical representations. 
As described in Section~\ref{sec:overall}, EVOLVE constructs two historical views during training: the normal historical view $H_T^n$ and the anomaly-aware historical view $H_T^a$. 
UALM adopts shared parameters for both views so that normal and anomaly-aware semantics are represented in a unified latent space.

For notation simplicity, we denote the input to UALM as a generic historical representation $H_T$, where $H_T \in \{H_T^n, H_T^a\}$ during training. 
UALM first parameterizes a latent Gaussian distribution:
\begin{equation}
    [\mu, \rho] = \mathrm{MLP}_{z}(H_T),
\end{equation}
where $\mu \in \mathbb{R}^{L_Z}$ denotes the latent mean and $\rho \in \mathbb{R}^{L_Z}$ is transformed into a positive standard deviation:
\begin{equation}
    \sigma = \mathrm{softplus}(\rho) + \epsilon_0,
\end{equation}
with a small constant $\epsilon_0$ for numerical stability. 
Then $K$ latent semantic hypotheses are sampled through reparameterization:
\begin{equation}
    Z^{(k)} = \mu + \eta^{(k)} \odot \sigma,
    \quad
    \eta^{(k)} \sim \mathcal{N}(0,I),
    \quad k=1,\dots,K.
\end{equation}
The sampled latent set is denoted as
\begin{equation}
    \mathbf{Z}_K = \{Z^{(k)}\}_{k=1}^{K}.
\end{equation}
When the input is $H_T^n$ or $H_T^a$, the corresponding latent sets are denoted as $\mathbf{Z}_K^n$ and $\mathbf{Z}_K^a$, respectively. 
Each latent sample represents one plausible interpretation of the historical semantics.

In addition to latent sampling, UALM estimates the evolution-mode distribution for each latent hypothesis. 
For each $Z^{(k)}$, UALM concatenates it with the historical representation and outputs a two-dimensional logit vector:
\begin{equation}
    W^{(k)} = \mathrm{MLP}_{w}([Z^{(k)}; H_T]),
    \quad W^{(k)} \in \mathbb{R}^{2}.
\end{equation}
The two entries of $W^{(k)}$ correspond to the smooth evolution hypothesis and the deviation-amplifying evolution hypothesis. 
We then apply softmax along this two-hypothesis dimension:
\begin{equation}
    p^{(k)} = 
    \mathrm{softmax}(W^{(k)})
    =
    [p_{\mathrm{smooth}}^{(k)}, p_{\mathrm{dev}}^{(k)}],
\end{equation}
where $p_{\mathrm{smooth}}^{(k)} + p_{\mathrm{dev}}^{(k)} = 1$. 
Importantly, the softmax is applied over the two evolution hypotheses for each latent sample, rather than across the $K$ sampled paths. 
For normal and anomaly-aware historical views, the corresponding evolution-mode distributions are denoted as $\mathbf{P}_K^n$ and $\mathbf{P}_K^a$.

UALM differs from generic VAE-style stochastic forecasting in both objective and usage. 
It is not designed to maximize the likelihood of future observations or to directly use latent variance as anomaly scores. 
Instead, UALM samples multiple semantic hypotheses and associates each hypothesis with an evolution-mode distribution. 
These hypotheses are decoded into multiple future trajectories, and their predictive disagreement is later used to estimate future anomaly risk. 
Therefore, the uncertainty modeled by UALM is task-specific and risk-oriented for anomaly prediction.

\subsubsection{Dual-Structured Decoder}

Given the latent hypotheses $\mathbf{Z}_K$ and their evolution-mode probabilities $\mathbf{P}_K$, the \emph{Dual-Structured Decoder (DSD)} decodes each latent hypothesis into a plausible future trajectory. 
Different from deterministic forecasting decoders, DSD implements two complementary evolution hypotheses through two decoding branches: a smooth evolution hypothesis and a deviation-amplifying evolution hypothesis. 
The former describes stable future evolution where deviations or perturbations are absorbed by the system, while the latter describes the amplification of local deviations into potential abnormal dynamics.

For notation clarity, we describe the decoding process for a single latent hypothesis $Z^{(k)}$, while all $K$ hypotheses are processed in parallel in practice.

\noindent \textbf{Shared latent transformation.}
Each latent hypothesis is first projected into the future hidden space:
\begin{equation}
    \tilde{H}_F^{(k)}
    =
    \mathrm{Transformer}
    \left(
    \mathrm{MLP}(Z^{(k)})
    \right),
\end{equation}
where $\tilde{H}_F^{(k)}$ denotes the shared future hidden representation. 
This shared transformation captures common temporal dependencies and provides the basis for the following two evolution hypotheses.

\noindent \textbf{Smooth evolution hypothesis.}
The smooth evolution hypothesis assumes that potential deviations or perturbations are absorbed by the system and the future dynamics remain regular. 
It applies a SmoothTransformer block to the shared future hidden representation. 
For notation simplicity, let $H \equiv \tilde{H}_F^{(k)}$. 
The smooth branch is computed as
\begin{equation}
    H' =
    \mathrm{LN}
    \left(
    H + \mathrm{MHA}(H,H,H)
    \right),
\end{equation}
\begin{equation}
    H'' =
    \mathrm{LN}
    \left(
    H' + \mathrm{FFN}(H')
    \right),
\end{equation}
\begin{equation}
    H_{\mathrm{smooth}}^{(k)}
    =
    H'' + \mathrm{DWConv}(H''),
\end{equation}
where $\mathrm{MHA}(\cdot)$ captures global temporal dependencies, $\mathrm{FFN}(\cdot)$ refines the hidden representation, and $\mathrm{DWConv}(\cdot)$ denotes depth-wise temporal convolution. 
The depth-wise convolution suppresses high-frequency fluctuations and encourages stable and coherent future evolution.

\noindent \textbf{Deviation-amplifying evolution hypothesis.}
The deviation-amplifying evolution hypothesis models the case where local deviations are enlarged and evolve into potential abnormal dynamics. 
It processes the shared future representation through a dilated convolution block followed by a contrastive enhancement block. 
Let $H \equiv \tilde{H}_F^{(k)}$. 
The dilated convolution block is formulated as
\begin{equation}
    H' =
    \mathrm{LN}
    \left(
    H + \mathrm{PWConv}(\mathrm{DConv}(H))
    \right),
\end{equation}
where $\mathrm{DConv}(\cdot)$ denotes depth-wise dilated temporal convolution and $\mathrm{PWConv}(\cdot)$ denotes point-wise convolution for channel mixing. 
This block enlarges the temporal receptive field to capture sparse and long-range abnormal patterns.

To emphasize irregular components, the contrastive enhancement block first extracts a smoothed representation:
\begin{equation}
    H_{\mathrm{low}} =
    \mathrm{SmoothConv}(H'),
\end{equation}
and then computes high-frequency residuals:
\begin{equation}
    H_{\mathrm{high}} =
    \mathrm{LN}
    \left(
    H' - H_{\mathrm{low}}
    \right).
\end{equation}
The final deviation-amplifying representation is obtained as
\begin{equation}
    H_{\mathrm{dev}}^{(k)}
    =
    H' + H_{\mathrm{high}}.
\end{equation}
This module does not generate perturbations by itself; rather, it models how already injected or latent deviations may be amplified during future evolution.

\noindent \textbf{Evolution-mode fusion.}
For each latent hypothesis, the two decoded representations are fused according to the evolution-mode probabilities produced by UALM:
\begin{equation}
    H_F^{(k)}
    =
    p_{\mathrm{smooth}}^{(k)} \cdot H_{\mathrm{smooth}}^{(k)}
    +
    p_{\mathrm{dev}}^{(k)} \cdot H_{\mathrm{dev}}^{(k)}.
\end{equation}
The fused future hidden representation is then projected to the observation space through a lightweight prediction head:
\begin{equation}
    \hat{X}_{T+1:T+F}^{(k)}
    =
    \mathrm{Head}(H_F^{(k)}).
\end{equation}
The set of decoded future trajectories is denoted as
\begin{equation}
    \mathbf{\hat{X}}_F
    =
    \{\hat{X}_{T+1:T+F}^{(k)}\}_{k=1}^{K}.
\end{equation}

During training, DSD is applied to both the normal historical view and the anomaly-aware historical view, producing $\mathbf{\hat{X}}_F^n$ and $\mathbf{\hat{X}}_F^a$, respectively. 
The normal view is encouraged to produce consistent future trajectories, while the anomaly-aware view is encouraged to expose deviation-sensitive future evolution. 
During inference, only the normal historical view is used, and future anomaly scores are computed from the predictive variance among $\mathbf{\hat{X}}_F^n$.

\subsection{Training Objective and Adversarial Optimization}
\label{sec:training}

EVOLVE is trained with a two-stage alternating optimization strategy. 
The first stage optimizes the predictor to learn discriminative future evolution from both the normal historical view and the anomaly-aware historical view. 
The second stage optimizes the perturbation generator to produce hard but plausible anomaly-aware views. 
This separation clarifies the roles of the predictor and the anomaly injection module during adversarial learning.

During training, EVOLVE constructs two historical views from the same input window: the normal historical view $H_T^n$ and the anomaly-aware historical view $H_T^a$. 
The normal view is expected to produce consistent future trajectories that match the observed future window, while the anomaly-aware view is expected to expose deviation-sensitive future evolution induced by variable-conditioned perturbations. 
Let
$\mathbf{\hat{X}}_F^n=\{(\hat{X}_F^n)^{(k)}\}_{k=1}^{K}$
and
$\mathbf{\hat{X}}_F^a=\{(\hat{X}_F^a)^{(k)}\}_{k=1}^{K}$
denote the $K$ decoded future trajectories from the normal and anomaly-aware historical views, respectively.

\noindent \textbf{Normal-view consistency and forecasting losses.}
For the normal historical view, multiple decoded future trajectories should be mutually consistent and close to the observed future window $X_{T+1:T+F}$. 
We define the predictive variance across the $K$ trajectories as
\begin{equation}
    \mathrm{Var}_K(\mathbf{\hat{X}}_F^n)
    =
    \frac{1}{FC}
    \sum_{t=T+1}^{T+F}
    \sum_{c=1}^{C}
    \mathrm{Var}
    \left(
    (\hat{x}_{t,c}^{n})^{(1)}, \dots, (\hat{x}_{t,c}^{n})^{(K)}
    \right).
\end{equation}
The normal-view consistency loss is then defined as
\begin{equation}
    \mathcal{L}_{\mathrm{con}}^{n}
    =
    \mathrm{Var}_K(\mathbf{\hat{X}}_F^n).
\end{equation}
Meanwhile, the mean prediction should reconstruct the observed future:
\begin{equation}
    \mathcal{L}_{\mathrm{rec}}^{n}
    =
    \mathrm{MSE}
    \left(
    \frac{1}{K}\sum_{k=1}^{K}(\hat{X}_F^n)^{(k)},
    X_{T+1:T+F}
    \right).
\end{equation}
The normal-view objective is
\begin{equation}
    \mathcal{L}_{n}
    =
    \mathcal{L}_{\mathrm{con}}^{n}
    +
    \mathcal{L}_{\mathrm{rec}}^{n}.
\end{equation}

\noindent \textbf{Anomaly-aware diversity and deviation losses.}
For the anomaly-aware historical view, the injected perturbation should induce futures that are distinguishable from stable normal evolution. 
Therefore, the decoded trajectories are encouraged to have sufficiently large predictive diversity and deviation from the observed future. 
Instead of reusing the injection strength symbol, we define a separate abnormal margin $m$:
\begin{equation}
    m = \lambda_m \cdot \mathrm{sg}(\gamma^2),
\end{equation}
where $\gamma$ is the injection strength in Eq.~\eqref{eq:adaptive_injection}, $\lambda_m$ is a scaling coefficient, and $\mathrm{sg}(\cdot)$ denotes the stop-gradient operation. 
Detaching the margin prevents trivial solutions where the model reduces the margin to satisfy the loss.

The anomaly-aware diversity loss is defined as
\begin{equation}
    \mathcal{L}_{\mathrm{div}}^{a}
    =
    \mathrm{ReLU}
    \left(
    m -
    \mathrm{Var}_K(\mathbf{\hat{X}}_F^a)
    \right),
\end{equation}
where $\mathrm{Var}_K(\mathbf{\hat{X}}_F^a)$ is computed in the same way as the normal-view predictive variance. 
The anomaly-aware deviation loss is defined as
\begin{equation}
    \mathcal{L}_{\mathrm{dev}}^{a}
    =
    \mathrm{ReLU}
    \left(
    m -
    \mathrm{MSE}
    \left(
    \frac{1}{K}\sum_{k=1}^{K}(\hat{X}_F^a)^{(k)},
    X_{T+1:T+F}
    \right)
    \right).
\end{equation}
The anomaly-aware objective is
\begin{equation}
    \mathcal{L}_{a}
    =
    \mathcal{L}_{\mathrm{div}}^{a}
    +
    \mathcal{L}_{\mathrm{dev}}^{a}.
\end{equation}

\noindent \textbf{Evolution-mode alignment loss.}
UALM outputs a two-dimensional evolution-mode logit vector $W^{(k)}$ for each latent hypothesis, corresponding to the smooth evolution hypothesis and the deviation-amplifying evolution hypothesis. 
Let $W_n^{(k)}$ and $W_a^{(k)}$ denote the logits produced from the normal and anomaly-aware historical views, respectively. 
The normal view is encouraged to align with the smooth evolution hypothesis, while the anomaly-aware view is encouraged to align with the deviation-amplifying evolution hypothesis:
\begin{equation}
    \mathcal{L}_{\mathrm{align}}
    =
    \frac{1}{K}
    \sum_{k=1}^{K}
    \left[
    \mathrm{CE}(W_n^{(k)}, 0)
    +
    \mathrm{CE}(W_a^{(k)}, 1)
    \right],
\end{equation}
where class $0$ denotes the smooth evolution hypothesis and class $1$ denotes the deviation-amplifying evolution hypothesis. 
Equivalently, the one-hot targets are $[1,0]$ for the normal view and $[0,1]$ for the anomaly-aware view.

\noindent \textbf{Predictor optimization.}
In the predictor optimization stage, the anomaly injection module is fixed, and the predictor, including the encoder, UALM, DSD, and prediction head, is optimized by minimizing
\begin{equation}
    \mathcal{L}_{\mathrm{pred}}
    =
    \mathcal{L}_{n}
    +
    \mathcal{L}_{a}
    +
    \mathcal{L}_{\mathrm{align}}.
\end{equation}
This objective encourages the normal historical view to produce consistent and accurate future trajectories, while encouraging the anomaly-aware historical view to reveal deviation-sensitive future evolution.

\noindent \textbf{Perturbation generator optimization.}
In the perturbation generator optimization stage, the predictor is fixed and only the anomaly injection module is updated. 
The goal is to prevent the generator from producing overly obvious or trivial perturbations. 
Instead, the generator is encouraged to construct hard but plausible anomaly-aware views whose evolution-mode distributions are close to those of normal views.

For each latent hypothesis, we first compute the evolution-mode probabilities:
\begin{equation}
    p_n^{(k)} = \mathrm{softmax}(W_n^{(k)}),
    \quad
    p_a^{(k)} = \mathrm{softmax}(W_a^{(k)}).
\end{equation}
Here, softmax is applied along the two-hypothesis dimension of each logit vector, rather than across the $K$ sampled future paths. 
The adversarial loss for the perturbation generator is then defined as
\begin{equation}
    \mathcal{L}_{\mathrm{adv}}^{G}
    =
    \frac{1}{K}
    \sum_{k=1}^{K}
    \mathrm{KL}
    \left(
    p_a^{(k)}
    \parallel
    \mathrm{sg}(p_n^{(k)})
    \right),
\end{equation}
where $\mathrm{sg}(\cdot)$ stops gradients from flowing into the predictor. 
The generator is optimized by minimizing
\begin{equation}
    \mathcal{L}_{G}
    =
    \lambda_{\mathrm{adv}}
    \mathcal{L}_{\mathrm{adv}}^{G},
\end{equation}
where $\lambda_{\mathrm{adv}}$ controls the strength of adversarial perturbation learning.

The discriminator is not implemented as an additional binary classifier. 
Instead, the predictor's evolution-mode estimator plays the role of discrimination by assigning different smooth/deviation-amplifying probabilities to normal and anomaly-aware historical views.

The overall training alternates between predictor optimization and perturbation generator optimization. 
Through this adversarial process, EVOLVE learns variable-conditioned perturbations that are challenging enough to improve anomaly sensitivity, while maintaining plausible evolution patterns for unsupervised anomaly prediction.

\subsection{Inference and Anomaly Scoring}
\label{sec:inference}

During inference, EVOLVE estimates future anomaly risk using only the normal historical view. 
Anomaly-aware perturbation injection, Anomaly Pattern Analysis (APA), and dynamic prototype updating are disabled. 
Therefore, inference does not use future observations and does not construct anomaly-aware historical views, avoiding any test-time access to future information.

Given a historical window $X_{1:T}$, EVOLVE first applies instance normalization, embedding, and the time series encoder to obtain the normal historical representation $H_T^n$. 
The representation is then fed into UALM, which samples $K$ latent semantic hypotheses:
\begin{equation}
    \mathbf{Z}_K^n = \{(Z^n)^{(k)}\}_{k=1}^{K}.
\end{equation}
For each latent hypothesis, UALM also produces an evolution-mode probability over the smooth evolution hypothesis and the deviation-amplifying evolution hypothesis. 
These latent hypotheses are decoded by DSD to generate $K$ plausible future trajectories:
\begin{equation}
    \mathbf{\hat{X}}_F^n =
    \{(\hat{X}_F^n)^{(k)}\}_{k=1}^{K}.
\end{equation}

\begin{table*}[ht!]
\centering
\caption{Anomaly prediction performance of all models on six datasets}
\label{tab:34}

\renewcommand{\arraystretch}{1.15}
\resizebox{\textwidth}{!}{
\begin{tabular}{c|ccc|ccc|ccc|ccc|ccc|ccc}
\toprule

\multicolumn{1}{c|}{\textbf{Dataset}} & 
\multicolumn{3}{c|}{\textbf{MSL}} & 
\multicolumn{3}{c|}{\textbf{PSM}} & 
\multicolumn{3}{c|}{\textbf{SMD}} & 
\multicolumn{3}{c|}{\textbf{SWAN}} & 
\multicolumn{3}{c|}{\textbf{GECCO}} &
\multicolumn{3}{c}{\textbf{Genisis}} \\
\midrule

\multicolumn{1}{c|}{\textbf{Metric}}
& Aff-F1 & V-ROC & V-PR 
& Aff-F1 & V-ROC & V-PR 
& Aff-F1 & V-ROC & V-PR 
& Aff-F1 & V-ROC & V-PR 
& Aff-F1 & V-ROC & V-PR 
& Aff-F1 & V-ROC & V-PR \\

\midrule

LODA          & 63.78 & 53.53 & 10.57 & 61.02 & 48.28 & 30.41 & 57.93 & 39.14 & 5.34 & 37.11 & 18.38 & 30.04 & 39.04 & 43.27 & 2.23 & 54.25 & 46.62 & 0.93 \\
IForest       & 66.59 & 53.31 & 9.17 & 65.25 & 50.17 & 31.86 & 58.69 & 39.91 & 3.53 & 30.00 & 27.45 & 32.54 & 46.49 & 39.86 & 1.93 & 61.25 & 50.06 & 1.41 \\
TranAD        & 65.72 & 52.90 & 16.77 & 64.01 & 43.13 & 34.31 & 63.43 & 42.55 & 4.53 & 51.30 & 37.07 & 30.44 & 41.13 & 22.76 & 1.66 & 63.26 & 64.33 & 0.94 \\
DCdetector    & 59.41 & 49.68 & 10.69 & 66.08 & 37.49 & 27.02 & 65.51 & 40.43 & 4.69 & 52.66 & 32.92 & 32.99 & 52.62 & 41.21 & 1.71 & 67.56 & 53.82 & 1.25 \\
ATransformer  & 54.57 & 48.66 & 13.81 & 53.20 & 43.56 & 30.09 & 47.70 & 47.28 & 4.08 & 40.99 & 27.51 & 30.84 & 49.58 & 43.54 & 0.84 & 45.26 & 42.20 & 0.35 \\
CATCH         & 68.13 & 54.89 & 17.39 & 66.78 & 45.62 & 32.55 & 64.28 & 46.58 & 5.70 & 55.77 & 35.00 & 34.73 & 61.05 & 47.49 & 2.33 & 70.77 & 61.60 & 1.24 \\
CrossAD       & \underline{68.35} & 52.19 & 17.41 & 68.16 & 44.70 & 28.95 & 64.84 & 45.52 & 5.09 & 52.41 & 34.06 & 35.58 & 60.52 & 41.28 & \underline{2.42} & 71.11 & 61.68 & 1.40 \\
\cmidrule{1-19}
PAD           & 62.91 & 57.42 & 19.82 & 58.72 & 38.54 & 31.15 & 62.95 & 46.16 & 4.79 & 52.72 & 38.32 & 34.10 & 58.14 & 46.12 & 1.40 & 60.47 & 54.92 & 0.65 \\
A2P           & 67.93 & 59.62 & 17.80 & \underline{67.46} & 48.26 & 35.92 & 65.54 & \underline{49.97} & 5.33 & \underline{58.85} & 39.05 & 36.31 & 63.71 & \underline{49.69} & 1.85 & 72.88 & 63.12 & 0.99 \\
IGCL          & 66.33 & \underline{63.07} & 20.95 & 63.79 & \underline{51.58} & 36.03 & 67.39 & 49.88 & \underline{6.28} & 55.11 & 35.20 & 34.88 & 60.04 & 45.14 & 2.11 & \underline{74.11} & \underline{70.19} & 1.23 \\
RED-F         & 64.06 & 60.63 & \underline{21.21} & 69.97 & 45.71 & 34.35 & \underline{67.94} & 42.20 & 6.20 & 58.47 & \underline{40.11} & 36.08 & 61.41 & 47.51 & 2.07 & 68.97 & 61.69 & 1.36 \\
TAP           & 67.39 & 62.17 & 19.34 & 67.29 & 49.15 & \underline{36.04} & 67.16 & 49.16 & 6.16 & 55.56 & 35.09 & \underline{37.73} & \underline{64.46} & 45.71 & 1.93 & 66.91 & 69.77 & \underline{1.76} \\
FCM           & 66.26 & 57.51 & 17.62 & 60.66 & 49.30 & 32.58 & 64.71 & 48.87 & 5.87 & 55.94 & 39.49 & 35.62 & 61.30 & 46.99 & 2.13 & 60.86 & 59.99 & 1.43 \\
\cmidrule{1-19}
\textbf{EVOLVE} & \textbf{72.70} & \textbf{70.23} & \textbf{23.81} & \textbf{73.30} & \textbf{52.70} & \textbf{36.12} & \textbf{72.88} & \textbf{57.77} & \textbf{ 6.93} & \textbf{65.08} & \textbf{42.47} & \textbf{37.81} & \textbf{65.81} & \textbf{65.56} & \textbf{3.98} & \textbf{76.94} & \textbf{72.00} & \textbf{2.08} \\

\midrule
\bottomrule

\end{tabular}}
\end{table*}

Unlike training, the true future window $X_{T+1:T+F}$ is unavailable during inference, so reconstruction errors cannot be used for anomaly scoring. 
Instead, EVOLVE estimates anomaly risk from the predictive disagreement among the $K$ decoded future trajectories. 
Specifically, the anomaly score at each future time step is computed by the predictive variance defined in Eq.~\eqref{eq:anomaly_score}. 
A low predictive variance indicates that multiple plausible latent hypotheses lead to consistent future trajectories, suggesting stable future evolution. 
In contrast, a high predictive variance indicates that plausible future trajectories diverge, suggesting uncertain evolution and higher anomaly risk.

Finally, binary future anomaly labels are obtained by thresholding the anomaly scores:
\begin{equation}
    \hat{y}_t =
    \mathbb{I}(s_t \ge \tau),
    \quad t=T+1,\dots,T+F,
\end{equation}
where $\tau$ is the decision threshold and $\mathbb{I}(\cdot)$ denotes the indicator function.

\section{Experiments}

This section evaluates EVOLVE on six real-world multivariate time series benchmarks. 
We compare EVOLVE with both adapted anomaly detection methods and anomaly prediction methods, and analyze its effectiveness, efficiency, and component contributions. 
The implementation and supplemental experimental materials are publicly available online.\footnote{\url{https://github.com/jianxinjin038-sketch/Evolve}}

\subsection{Experimental Setup}

\subsubsection{Datasets and Protocol}

We evaluate EVOLVE on six real-world multivariate time series datasets: MSL~\cite{MSL}, PSM~\cite{PSM}, SMD~\cite{history_uncertainty_1}, SWAN~\cite{SWAN_GECCO}, GECCO~\cite{SWAN_GECCO}, and Genesis~\cite{Genesis}. 
These datasets cover diverse industrial and cyber-physical scenarios, including spacecraft telemetry, server machines, space weather monitoring, water treatment systems, and cyber-physical production systems. 
Following previous work~\cite{MSL_get_1}, only the first continuous segment of MSL is retained. 
Detailed dataset statistics, including the number of variables, training length, test length, and anomaly ratio, are provided in the supplemental materials.

For each dataset, we construct samples using a historical window of length $T$ and predict anomaly risks over a future horizon of length $F$. 
Unless otherwise specified, we set $T=96$ and evaluate four prediction horizons $F\in\{32,64,128,192\}$. 
During training, future observations are used only as self-supervised signals, while future anomaly labels are not used. 
During inference, only the historical window is available, and anomaly scores are computed from the predictive variance among decoded future trajectories. 
Missing values, if any, are handled by the standard preprocessing protocol used in prior time series anomaly detection benchmarks, and all input windows are normalized before being fed into the model.

\subsubsection{Baselines}

We compare EVOLVE with representative methods from two categories.

\textbf{Adapted anomaly detection methods.}
Following the adaptation protocols used in previous anomaly prediction studies~\cite{existing_AP_methods_1, existing_AP_methods_2}, we include widely used anomaly detection methods and adapt them to the anomaly prediction setting. 
These baselines include statistical and density-based methods, i.e., LODA~\cite{LODA} and IForest~\cite{IForest}; reconstruction- or forecasting-based deep models, i.e., TranAD~\cite{TranAD}, Anomaly Transformer~\cite{Atransformer}, CATCH~\cite{CATCH}, and CrossAD~\cite{CrossAD}; and contrastive learning-based models, i.e., DCdetector~\cite{channel_independent_2}.

\textbf{Anomaly prediction methods.}
We further compare EVOLVE with methods specifically designed for anomaly prediction, including PAD~\cite{AP_after_time_4}, A2P~\cite{A2P}, IGCL~\cite{IGCL}, RED-F~\cite{existing_AP_methods_4}, TAP~\cite{existing_AP_methods_1}, and FCM~\cite{existing_AP_methods_5}. 
These methods cover precursor-based, perturbation-based, contrastive forecasting, and future-context modeling paradigms for anomaly prediction. 
For all baselines, we use official implementations when available and follow the recommended configurations. 
When direct anomaly prediction outputs are unavailable, we adapt the methods following protocols established by previous AP studies.

\subsubsection{Evaluation Metrics}

We report Affiliation-F1 (Aff-F1)~\cite{AFF}, VUS-ROC, and VUS-PR~\cite{R_V, vus} as the primary metrics in the main tables, since anomaly prediction is highly imbalanced and temporally structured. 
Aff-F1 evaluates the temporal affiliation between predicted and ground-truth anomaly events, making it more suitable than point-wise F1 when predictions are slightly shifted in time. 
VUS-ROC and VUS-PR provide range-aware evaluation over different temporal ranges and thresholds, with VUS-PR being particularly informative under severe class imbalance.


\subsubsection{Implementation Details}
\label{sec:implementation}

All experiments are implemented in PyTorch and conducted on an NVIDIA GeForce RTX 3090 GPU with 24GB memory. 
We use the Adam optimizer~\cite{Adam} with a learning rate of $0.001$ for training. 
The random seed is fixed to $2026$ for reproducibility. 
Following the testing protocol proposed in TFB~\cite{TFB}, we disable the \texttt{drop\_last} operation during evaluation to avoid evaluation bias caused by inconsistent test batch sizes.

For EVOLVE, the size of the dynamic prototype memory is set to $100$. 
The number of latent hypotheses $K$, the number of clusters in APA, and the batch size are treated as hyperparameters. 
The initial batch size is set to $64$ and is halved when an out-of-memory error occurs, with a minimum batch size of $8$. 
The dynamic prototype bank is initialized as an empty FIFO queue and is updated only during training. 
During inference, anomaly-aware perturbation injection and dynamic prototype updating are disabled.

Unless otherwise specified, equal weights are assigned to the loss terms in the predictor objective. 
For all baseline methods, we use their official or open-source implementations whenever available, and follow the configurations recommended in the original papers. 
All methods are evaluated under the same hardware environment and testing protocol for fair comparison.

\subsection{Main Results}
\label{sec:main_results}

We conduct a systematic evaluation on six multivariate time series datasets under four prediction horizons, i.e., $F \in \{32,64,128,192\}$. 
The historical window is fixed as $T=96$ for all datasets and methods. 
For clarity, Table~\ref{tab:34} reports the average performance across the four prediction horizons, while full horizon-specific results are provided in the supplemental materials.

As shown in Table~\ref{tab:34}, EVOLVE achieves the best average performance on all six datasets across the three primary metrics, including Aff-F1, VUS-ROC, and VUS-PR. 
This demonstrates that EVOLVE can better localize future anomaly events and more reliably rank future anomaly risks, which is essential for early-warning systems in real-world applications.

Compared with adapted anomaly detection methods, EVOLVE shows consistent advantages across diverse datasets. 
Although methods such as CATCH, CrossAD, DCdetector, and Anomaly Transformer achieve competitive performance in some cases, they are primarily designed to detect anomalies from observed windows rather than predict future abnormal evolution. 
Thus, directly adapting them to AP can be insufficient when future anomaly risks depend on uncertain and heterogeneous evolution patterns.

Compared with anomaly prediction baselines, including PAD, A2P, IGCL, RED-F, TAP, and FCM, EVOLVE also achieves stronger performance. 
This indicates that existing precursor-based, perturbation-based, and future-context modeling methods still have limitations in capturing variable-conditioned anomaly responses and uncertain future evolution. 
In contrast, EVOLVE constructs anomaly-aware historical views through adaptive prototype retrieval and estimates anomaly risk from the predictive disagreement among multiple decoded future trajectories, enabling more reliable future anomaly prediction.

\subsection{Ablation Studies}
\label{sec:ablation}

We conduct ablation studies on all six datasets to evaluate the contribution of each major component in EVOLVE. 
Due to space constraints, Tables~\ref{tab:2}--\ref{tab:4} report representative results on MSL, SMD, and SWAN, while the complete six-dataset ablation results are provided in the supplemental materials. 
These three datasets cover different application domains and anomaly characteristics, including spacecraft telemetry, server-machine monitoring, and space weather monitoring, providing a compact yet diverse evaluation of the component contributions. 
The full results in the supplemental materials show consistent trends across all datasets.

Table~\ref{tab:2} studies the effectiveness of the dual-structured decoder. 
We compare four variants: removing both evolution hypotheses, using only the smooth evolution hypothesis, using only the deviation-amplifying evolution hypothesis, and using the full dual-structured decoder. 
The results show that using either hypothesis alone is insufficient to consistently achieve the best performance. 
The smooth evolution hypothesis helps model stable future dynamics, while the deviation-amplifying evolution hypothesis captures potential abnormal dynamics induced by local deviations. 
Combining them leads to the best overall performance, demonstrating that the two evolution hypotheses provide complementary information for multi-hypothesis future modeling.

\begin{table}[ht!]
\centering
\caption{ABLATION OF THE DUAL-STRUCTURED DECODER.}
\label{tab:2}

\renewcommand{\arraystretch}{1.15}
\resizebox{\columnwidth}{!}{
\begin{tabular}{cc|ccc|ccc|ccc}
\toprule

\multicolumn{2}{c|}{\textbf{Dataset}} & 
\multicolumn{3}{c|}{\textbf{MSL}} & 
\multicolumn{3}{c|}{\textbf{SMD}} & 
\multicolumn{3}{c}{\textbf{SWAN}} \\
\midrule

\multicolumn{1}{c}{\textbf{Smooth}} &
\multicolumn{1}{c|}{\textbf{Alert}}
& AFF-F1 & V-ROC & V-PR 
& AFF-F1 & V-ROC & V-PR 
& AFF-F1 & V-ROC & V-PR \\

\midrule

\xmark & \xmark & 70.21 & 65.41 & 20.88 & 72.65 & 54.65 &  6.85 & 63.40 & 40.02 & 35.96  \\
\cmark & \xmark & 68.75 & 64.56 & 20.55 & 70.97 & 56.40 &  6.76 & 63.36 & 39.26 & 35.16  \\
\xmark & \cmark & 68.78 & 63.59 & 18.55 & 72.05 & 55.97 &  6.73 & 61.10 & 42.08 & 36.52  \\
\cmark & \cmark & \textbf{72.70} & \textbf{70.23} & \textbf{23.81} & \textbf{72.88} & \textbf{57.77} & \textbf{6.93} & \textbf{65.08} & \textbf{42.47} & \textbf{37.81} \\

\midrule
\bottomrule

\end{tabular}}
\end{table}

Table~\ref{tab:3} evaluates the contribution of key components in EVOLVE. 
Removing adaptive anomaly injection degrades performance, indicating that variable-conditioned perturbation modeling is important for capturing heterogeneous anomaly responses. 
Removing Anomaly Pattern Analysis also leads to performance degradation, showing that dynamically updated residual-based prototypes provide useful context-dependent anomaly patterns beyond fixed priors. 
When Uncertainty-Aware Latent Modeling is removed, the performance decreases, verifying the importance of sampling multiple latent semantic hypotheses for uncertain future evolution. 
Removing adversarial optimization further weakens the model, suggesting that hard but plausible perturbations help improve the discrimination between stable and deviation-sensitive future evolution. 
Overall, the full EVOLVE model achieves the best or competitive results, confirming the complementary roles of these components.

\begin{table}[ht!]
\centering
\caption{ABLATION OF KEY COMPONENTS.}
\label{tab:3}

\renewcommand{\arraystretch}{1.15}
\resizebox{\columnwidth}{!}{
\begin{tabular}{c|ccc|ccc|ccc}
\toprule

\multicolumn{1}{c|}{\textbf{Dataset}} & 
\multicolumn{3}{c|}{\textbf{MSL}} & 
\multicolumn{3}{c|}{\textbf{SMD}} & 
\multicolumn{3}{c}{\textbf{SWAN}} \\
\midrule

\multicolumn{1}{c|}{\textbf{Metric}}
& AFF-F1 & V-ROC & V-PR 
& AFF-F1 & V-ROC & V-PR 
& AFF-F1 & V-ROC & V-PR \\

\midrule

w/o Dynamically Adaptive Anomalies & 70.51 & 64.63 & 21.24 & 70.57 & 51.31 &  6.26 & 59.50 & 41.58 & 36.55 \\
w/o Anomaly Pattern Analysis & 70.57 & 65.99 & 20.98 & 71.54 & 56.85 &  6.62 & 60.21 & 40.62 & 36.06 \\
w/o Adaptive Latent Modeling & 69.55 & 66.09 & 21.55 & 71.07 & 56.03 & 6.65 & 61.92 & 42.12 & 36.51 \\
w/o Adversarial Anomaly Generation & 70.21 & 63.35 & 20.84 & 70.78 & 57.06 & 6.50 & 62.68 & 39.94 & 36.06 \\
\textbf{EVOLVE} & \textbf{72.70} & \textbf{70.23} & \textbf{23.81} & \textbf{72.88} & \textbf{57.77} & \textbf{6.93} & \textbf{65.08} & \textbf{42.47} & \textbf{37.81} \\

\midrule
\bottomrule

\end{tabular}}
\end{table}

\begin{table}[ht!]
\centering
\caption{ABLATION OF TRAINING LOSSES.}
\label{tab:4}

\renewcommand{\arraystretch}{1.15}
\resizebox{\columnwidth}{!}{
\begin{tabular}{c|ccc|ccc|ccc}
\toprule

\multicolumn{1}{c|}{\textbf{Dataset}} & 
\multicolumn{3}{c|}{\textbf{MSL}} & 
\multicolumn{3}{c|}{\textbf{SMD}} & 
\multicolumn{3}{c}{\textbf{SWAN}} \\
\midrule

\multicolumn{1}{c|}{\textbf{Metric}}
& AFF-F1 & V-ROC & V-PR 
& AFF-F1 & V-ROC & V-PR 
& AFF-F1 & V-ROC & V-PR \\

\midrule

w/o MSE Loss & 70.44 & 65.66 & 21.01 & 71.96 & 55.68 & 6.61 & 64.88 & 40.47 & 35.67 \\
w/o VAR Loss & 69.54 & 64.51 & 20.17 & 70.90 & 56.23 & 6.36 & 60.80 & 41.98 & 35.24 \\
w/o Weights Loss & 69.03 & 66.05 & 20.11 & 70.55 & 54.45 & 6.19 & 62.52 & 41.25 & 36.60 \\
\textbf{EVOLVE} & \textbf{72.70} & \textbf{70.23} & \textbf{23.81} & \textbf{72.88} & \textbf{57.77} & \textbf{6.93} & \textbf{65.08} & \textbf{42.47} & \textbf{37.81} \\

\midrule
\bottomrule

\end{tabular}}
\end{table}

Table~\ref{tab:4} analyzes the effect of different training objectives. 
Removing the MSE-based objectives weakens the alignment between decoded trajectories and observed future dynamics. 
Removing the variance-based objectives reduces the model's ability to distinguish stable future evolution from uncertain or deviation-sensitive evolution. 
Removing the evolution-mode alignment loss also leads to performance degradation, indicating that explicitly aligning normal historical views with smooth evolution and anomaly-aware historical views with deviation-amplifying evolution is beneficial. 
These results demonstrate that the proposed training objectives jointly contribute to accurate, diverse, and semantically consistent anomaly prediction.





\subsection{Parameter Sensitivity}
\label{sec:parameter_sensitivity}

We further analyze the sensitivity of EVOLVE to several key hyperparameters, including the number of latent hypotheses $K$, the latent dimension, the number of anomaly prototype clusters $n_{\mathrm{clusters}}$, and the adversarial loss weight $\lambda_{\mathrm{adv}}$. 
Experiments are conducted on MSL, SMD, and SWAN using VUS-PR as the evaluation metric.

\begin{figure}[t]
  \centering
  \includegraphics[width=\columnwidth]{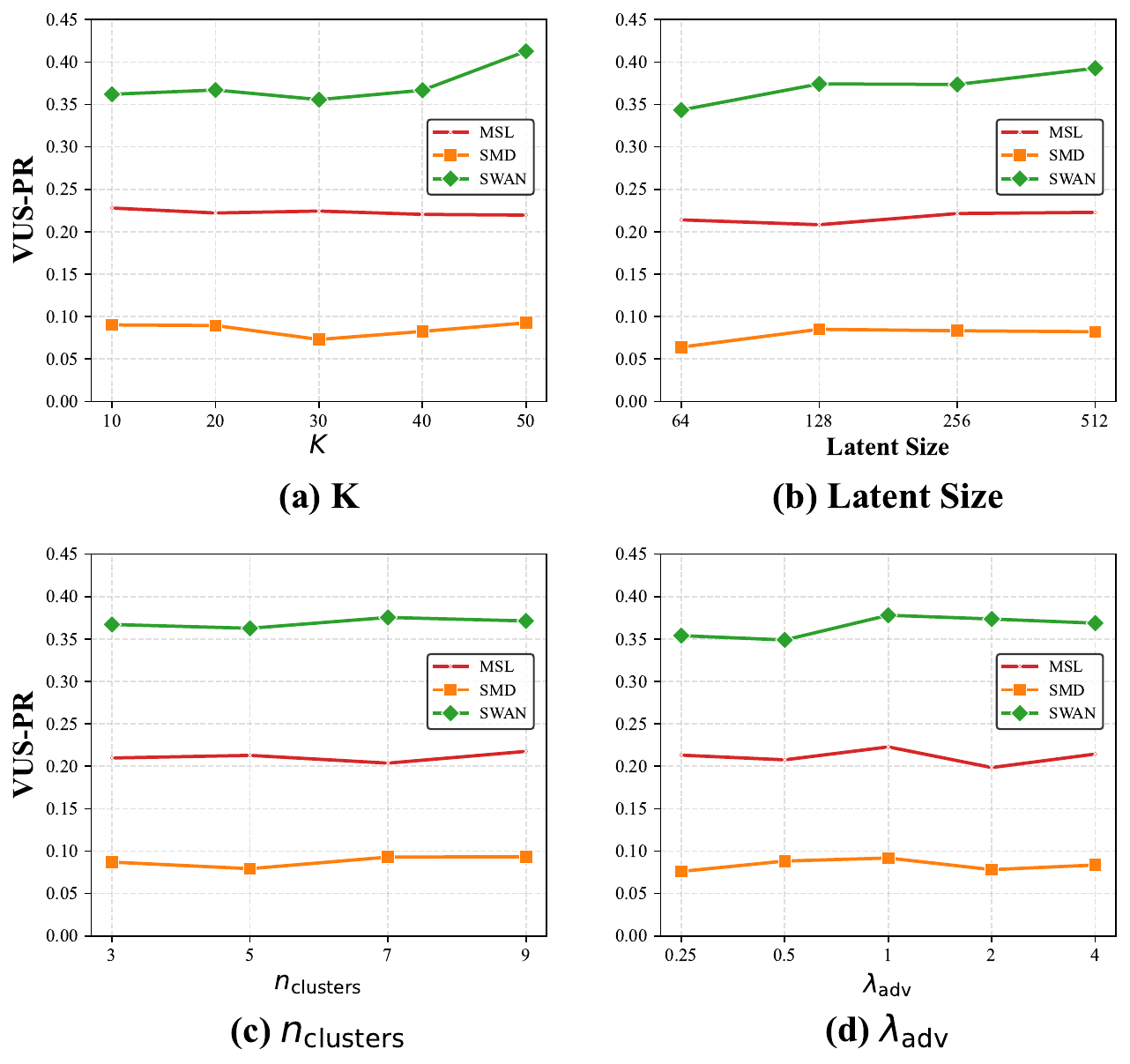}
  \caption{Parameter sensitivity analysis of EVOLVE on MSL, SMD, and SWAN using VUS-PR.}
  \label{fig:Parameter_Sensitivity}
\end{figure}

As shown in Figure~\ref{fig:Parameter_Sensitivity}, EVOLVE maintains relatively stable performance across different hyperparameter settings. 
Varying $K$ does not lead to drastic performance changes, indicating that a moderate number of latent hypotheses is sufficient to capture future uncertainty. 
The model is also not overly sensitive to the latent dimension or the number of anomaly prototype clusters, suggesting that UALM and APA remain effective under different representation capacities and clustering granularities. 
Similarly, changing $\lambda_{\mathrm{adv}}$ only causes limited fluctuations, indicating that adversarial perturbation learning improves anomaly-aware representation learning without destabilizing training.

Overall, these results show that EVOLVE is not highly sensitive to the tested hyperparameters and can maintain stable anomaly prediction performance without extensive dataset-specific tuning.

\subsection{Efficiency and Scalability Analysis}
\label{sec:efficiency}

We further evaluate the computational efficiency and scalability of EVOLVE on the MSL dataset, with the historical window fixed as $T=96$ and the prediction horizon fixed as $F=32$. 
Table~\ref{tab:5} compares EVOLVE with representative anomaly prediction baselines in terms of training time, inference time, memory consumption, and prediction performance measured by VUS-ROC.

\begin{table}[t]
\centering
\caption{Efficiency and effectiveness comparison on MSL.} 
\label{tab:5}

\renewcommand{\arraystretch}{1.15}
\resizebox{\columnwidth}{!}{
\begin{tabular}{c|c|c|c|c}
\toprule
\textbf{Method} & 
\textbf{VUS-ROC} & 
\textbf{Training Time} & 
\textbf{Inference Time} & 
\textbf{Memory Used} \\
\midrule

IGCL   & 66.74 & 14.09s & 0.19s & 34.12MB \\
PAD    & 60.79 & 99.20s & 0.86s & 281.13MB \\
A2P    & 62.01 & 34.87s & 1.36s & 559.77MB \\
RED-F  & 54.96 & 78.54s & 2.15s & 567.22MB \\
TAP    & 71.04 & 97.33s & 1.85s & 496.63MB \\
FCM    & 57.04 & 75.56s & 1.97s & 443.53MB \\
\textbf{EVOLVE} & \textbf{76.17} & 67.21s & 0.37s & 145.98MB \\

\bottomrule
\end{tabular}}
\vspace{-4pt}
\end{table}

As shown in Table~\ref{tab:5}, EVOLVE achieves the best VUS-ROC among all compared methods, outperforming the strongest baseline TAP by $5.13$ absolute points. 
Although EVOLVE requires more training time than lightweight methods such as IGCL and A2P, its training cost is lower than PAD, RED-F, TAP, and FCM. 
More importantly, EVOLVE maintains low inference latency and moderate memory consumption, with substantially faster inference than PAD, A2P, RED-F, TAP, and FCM. 
These results indicate that EVOLVE provides a favorable trade-off between predictive performance and computational cost.

The additional training cost mainly comes from two-stage alternating optimization, adaptive prototype updating, and adversarial perturbation learning. 
During inference, anomaly-aware perturbation injection and dynamic prototype updating are disabled, so the additional inference overhead remains limited. 
A backbone-controlled comparison is provided in the supplemental materials, showing that the improvement of EVOLVE does not solely come from the encoder backbone.

\begin{table}[t]
\centering
\caption{Scalability analysis with respect to the anomaly prototype memory size $N_{\mathrm{mem}}$.}
\label{tab:6}
\renewcommand{\arraystretch}{1.00}
\resizebox{0.8\columnwidth}{!}
{
\begin{tabular}{c|c|c|c}
\toprule
\textbf{$N_{\mathrm{mem}}$} & 
\textbf{Training Time} & 
\textbf{Memory Used} & 
\textbf{VUS-ROC} \\
\midrule

50            & 65.57s & 135MB & 75.64 \\
100           & 67.21s & 146MB & 76.17 \\
200           & 71.56s & 160MB & 77.81 \\
300           & 73.43s & 176MB & 76.10 \\
400           & 76.20s & 205MB & 75.54 \\
500           & 77.99s & 229MB & 74.78 \\

\bottomrule
\end{tabular}}
\vspace{-6pt}
\end{table}

Table~\ref{tab:6} analyzes the impact of the maximum size $N_{\mathrm{mem}}$ of the anomaly prototype memory. 
Unless otherwise specified, we set the default memory size to $N_{\mathrm{mem}}=100$ in all main experiments. 
As $N_{\mathrm{mem}}$ increases, training time and memory consumption grow gradually, while VUS-ROC remains relatively stable. 
This suggests that EVOLVE is not overly sensitive to memory capacity and can scale to larger prototype memories without causing excessive computational burden. 
The results also indicate that a compact memory is sufficient to provide useful anomaly prototypes for adaptive perturbation modeling.

\subsection{Visualization of Predictive Anomaly Scores}
\label{sec:visualization}

\begin{figure}[t]
  \centering
  \includegraphics[width=\columnwidth]{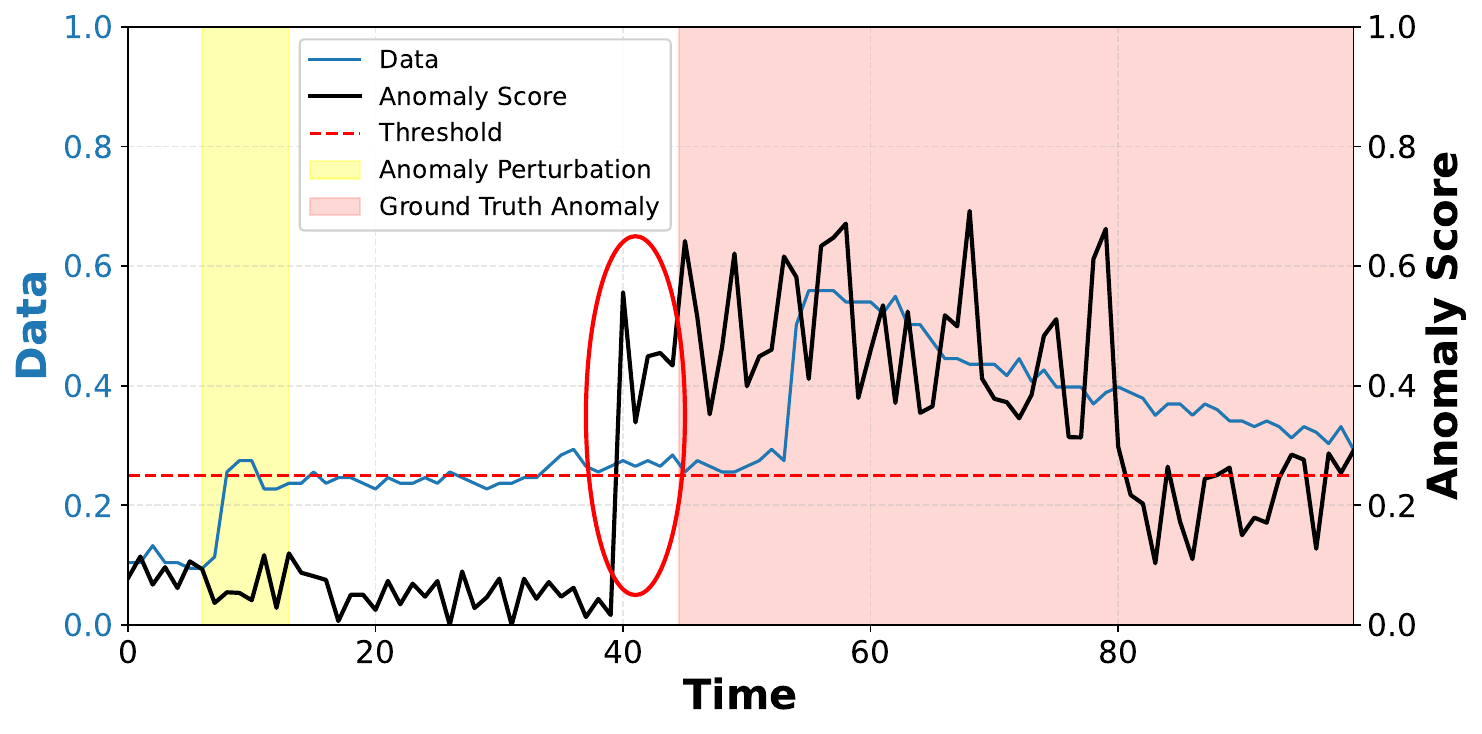}
  \caption{Visualization of predictive anomaly scores on the SMD dataset.} 
  \label{fig:anomaly_visualization}
  \vspace{-5pt}
\end{figure}

To further illustrate the behavior of the proposed anomaly scoring criterion, Figure~\ref{fig:anomaly_visualization} visualizes one variable from the SMD dataset together with the predictive variance-based anomaly scores produced by EVOLVE. 
A mild disturbance appears around $t\approx10$, while the anomaly score remains relatively stable. 
This indicates that EVOLVE does not simply react to instantaneous fluctuations in the observed history.

More importantly, the anomaly score rises sharply around $t\approx40$, before the ground-truth anomaly interval begins at approximately $t\approx45$. 
This early increase shows that the proposed scoring criterion can capture future anomaly risk before the abnormal segment fully manifests. 
After the anomaly develops, the anomaly score remains consistently high, reflecting sustained predictive uncertainty during abnormal evolution.

Overall, the visualization demonstrates the predictive nature of EVOLVE. 
Instead of relying on post-hoc reconstruction errors from observed anomalies, EVOLVE estimates future anomaly risk from the predictive disagreement among multiple decoded future trajectories, enabling early warning before anomalies become fully observable.

\section{Conclusion}
\vspace{-1pt}
In this paper, we propose EVOLVE, a unified framework for unsupervised multivariate time series anomaly prediction. 
EVOLVE models anomaly prediction through variable-conditioned perturbation modeling and uncertainty-aware multi-hypothesis future reasoning. 
To capture heterogeneous anomaly responses, EVOLVE introduces an adaptive anomaly prototype memory that combines fixed anomaly priors with dynamically updated prototypes, and retrieves context-specific perturbations to construct anomaly-aware historical views. 
To model uncertain future evolution, EVOLVE samples multiple latent semantic hypotheses and decodes them through smooth and deviation-amplifying evolution hypotheses, using the predictive disagreement among decoded future trajectories to estimate future anomaly risk. 
Extensive experiments on six real-world benchmark datasets demonstrate that EVOLVE outperforms state-of-the-art anomaly prediction baselines and adapted anomaly detection baselines, showing its effectiveness for early warning and proactive risk mitigation. 


\balance
\bibliographystyle{IEEEtran}
\bibliography{sample-base}

\end{document}